\documentclass[11pt,a4paper]{article}
\usepackage{times,latexsym}
\usepackage{url}
\usepackage[T1]{fontenc}
\usepackage{graphicx}
\usepackage{bm}
\graphicspath{ {./images/} }

\usepackage[acceptedWithA]{tacl2021v1}

\usepackage{booktabs}
\usepackage{xspace, mfirstuc, tabulary}

\usepackage{paralist, tabularx}
\usepackage{amsmath, amsfonts}

\newif\iftaclinstructions
\taclinstructionsfalse 
\iftaclinstructions
\renewcommand{\confidential}{}
\renewcommand{\anonsubtext}{(No author info supplied here, for consistency with
TACL-submission anonymization requirements)}
\newcommand{\instr}
\fi

\iftaclpubformat 

\else

\fi

\title{Distilling Multilingual Vision–Language Models: \\ When Smaller Models Stay Multilingual}

\author{
Sukrit Sriratanawilai\textsuperscript{$\heartsuit$}, 
Jhayahgrit Thongwat\textsuperscript{$\heartsuit$},
Romrawin Chumpu\textsuperscript{$\heartsuit$}, \\
\textbf{Patomporn Payoungkhamdee}\textsuperscript{$\heartsuit$},  
\textbf{Sarana Nutanong}\textsuperscript{$\heartsuit$},
\textbf{Peerat Limkonchotiwat}\textsuperscript{$\spadesuit$}\\
  \textsuperscript{$\heartsuit$}VISTEC,
  \textsuperscript{$\spadesuit$}AI Singapore 
   \\
  \texttt{sukrit.s\_s19@vistec.ac.th}, \texttt{peerat@aisingapore.org}
  }

\date{}

\begin{document}
\maketitle
\begin{abstract}

Vision–language models (VLMs) exhibit uneven performance across languages, a problem that is often exacerbated when the model size is reduced.
While Knowledge distillation (KD) demonstrates promising results in transferring knowledge from larger to smaller VLMs, applying KD in multilingualism is an underexplored area.  
This paper presents a controlled empirical study of KD behavior across five distillation approaches, isolating their effects on cross-lingual representation consistency and downstream performance stability under model compression.
We study five distillation formulations across CLIP and SigLIP2, and evaluate them on in-domain retrieval and out-of-domain visual QA. 
We find that some configurations preserve or even improve multilingual retrieval robustness despite halving model size, but others fail to maintain cross-task stability, exposing design-sensitive trade-offs that aggregate accuracy alone does not reveal.

\end{abstract}

\section{Introduction}

Vision–language models (VLMs) have become the dominant paradigm for joint visual–textual representation learning~\cite{clip,siglip}.
One prominent approach to achieving performance gains is to utilize a large-scale multilingual corpus~\cite{siglip2}.
This practice results in a reliance on massive encoder models.

VLM architectures vary widely in scale, from small models such as CLIP~\cite{clip}, FILIP~\cite{filip}, ALIGN~\cite{align2021} to large models like SigLIP~\cite{siglip}, SigLIP2~\cite{siglip2}, CoCa~\cite{coca}
Notably, a single text encoder can account for more than half of the total model size (e.g., 565M of 881M parameters, or 64\%, in SigLIP2-L/16).
In contexts where compute is inherently bounded, models at this scale are not merely inefficient but unusable~\cite{minilm, mobilebert, mobileformer}.

Smaller models present both benefits and challenges. 
Table~\ref{tab:compression_tradeoffs} reports preliminary results that illustrate this trade-off. 
Compressing SigLIP2-L/16 yields a substantial inference speedup but also degrades performance. 
Crucially, we observe degradation in both the image-to-text (I2T) retrieval and downstream CVQA accuracy. 
In particular, cross-lingual retrieval on the Multi30K dataset decreases by 12.58 points (from 73.43 to 60.85) for I2T and by 4.36 points (from 32.88 to 28.52) for CVQA when the number of parameters is reduced from 881M to 433M using the feature distillation (FD) method proposed by \citet{multilingualclip}.

\begin{table}[htbp]
\vspace{-2mm}
  \small
  \centering
  \setlength{\tabcolsep}{4pt}
  \scalebox{0.9}{
  \renewcommand{\arraystretch}{1}
  \begin{tabular}{
    >{\centering\arraybackslash}p{0.26\linewidth}   
    >{\centering\arraybackslash}p{0.25\linewidth}   
    >{\centering\arraybackslash}p{0.20\linewidth}   
    >{\centering\arraybackslash}p{0.20\linewidth}   
  }
    \toprule
    \textbf{Params (M)} & \textbf{Inference speed} & \textbf{Recall@1 I2T} & \textbf{CVQA Accuracy}\\
    \midrule
    881 (Teacher) & $1\times$ & 73.43 & 32.88 \\
    593 (Student) & $\sim3.5\times$ & 70.90 & 30.30 \\
    450 (Student) & $\sim6.1\times$ & 63.38 & 28.60 \\
    433 (Student) & $\sim9.7\times$ & 60.85 & 28.52 \\

    \bottomrule
  \end{tabular}}
  \vspace{-1mm}
  \caption{Effect of compressing SigLIP2-L/16 on inference speed, recall@1 (Multi30K), and accuracy (CVQA). While model size reduction yields a substantial speedup, the performance of downstream tasks decreases as the model size is reduced.}
  \vspace{-3mm}
  \label{tab:compression_tradeoffs}
\end{table}

This challenge becomes even more pronounced in multilingual settings.
In such environments, KD must also maintain multilingual consistency in cross-modal representations while reducing model size.
While KD receives significant attention~\cite{distillbert,minilm,multilingualclkd,clip-kd}, multilingual consistency is rarely an explicit consideration in existing KD objectives.
This consideration is crucial for enabling the efficient deployment of such models in linguistically diverse environments.

In this paper, we tackle this problem through the following two research questions:
\begin{compactitem}
    \item \textbf{RQ1:} Across different knowledge distillation strategies, which ones best preserve multilingual retrieval performance?

    \item \textbf{RQ2:} What impact does the knowledge distillation have on cross-lingual efficiency, latent structure, and retrieval robustness?

\end{compactitem}

In particular, we focus on investigating knowledge distillation in a multilingual vision-language model setting, specifically examining how knowledge can be effectively transferred from vision-language foundation models to multilingual encoder models.
To address \textbf{RQ1}, we investigate knowledge distillation techniques to improve the performance of multilingual downstream tasks, where we use five techniques covering major KD developments, as well as cross-lingual knowledge transfer solutions.
Then, we analyze the performance gap when transferring knowledge from VLM text representation to small multilingual encoders, such as XLM-R\textsubscript{Base}~\cite{xlmr}, DistilBERT~\cite{distillbert}, and MiniLM~\cite{minilm}.
To answer \textbf{RQ2}, we analyze the trade-offs involved, identifying when model size reduction preserves multilingual retrieval quality and when it leads to a performance decrease in some languages. 
We also examine which tasks, such as ranking and clustering, enable smaller models to remain competitive with, or even outperform, larger models.


We evaluate knowledge distillation across two representative teacher models with contrasting multilingual properties: CLIP-ViT-L/14, an English-centric model, and SigLIP2-L/16, a natively multilingual VLM. 
Our experiments cover both in-domain (image–text retrieval) and out-of-domain (visual question answering) benchmarks, allowing us to examine not only cross-lingual alignment preservation but also generalization beyond the teacher’s training distribution.
Our findings indicate that performance retention after compression is indeed achievable, but only under the right distillation configurations, with learning objectives playing a particularly critical role. 
Moreover, the effect is strongly task-dependent: while some settings suffer degradation, tasks such as multilingual reranking and clustering prove more resilient, with smaller models in some cases performing competitively with larger ones.

Our contributions are as follows:
\begin{compactitem}
    \item We present the first systematic comparative study of knowledge distillation strategies in multilingual vision–language models, characterizing how different design choices influence cross-lingual alignment and performance under compression.

    \item We analyze trade-offs, showing when size reduction preserves multilingual retrieval quality, when some languages perform poorly, and which tasks (e.g., ranking, clustering) allow smaller models to remain competitive with or outperform larger models.

\end{compactitem}

\section{Related Work}
\subsection{Multilingual Vision-language Model Training}

Recently, many works~\cite{clip, align2021, blip2023} have researched multimodality (vision-text encoder) by aggregating ViT~\cite{ViT} and text encoder~\cite{bert} and training the model using vision-text datasets (i.e., CC-12M~\cite{cc12m} or LAION~\cite{laion}).
The training objective of this kind of work is to maximize the similarity of text and image pairs, while minimizing the similarity of irrelevant pairs using contrastive learning~\cite{clip}.
This technique has been proven to be a robust training technique to achieve a strong vision-text encoder model on retrieval~\cite{vlmsretrievalenhancedcontrastivevisiontextmodels} or VQA~\cite{vlmskant2021contrast,vlmsvqaparelli2023clip}.
However, these previous works only experimented in English, while the multilingual capability might not have been well established.

Researchers extend the CLIP method to support multilingualism by aligning CLIP’s representations with those of multilingual texts in the same embedding space. 
Multilingual CLIP~\cite{multilingualclip} uses knowledge distillation from CLIP text’s encoder as teacher and multilingual encoder as student by using text pairs that are translated by machine translation as training data. 
mCLIP~\cite{mclip} aligns English text representation between CLIP’s encoder and multilingual encoder with Triangle cross-modal Knowledge Distillation loss.   
Recently, SIGLIP2~\cite{siglip2} has been proposed as a multilingual and multimodal foundation model trained from scratch on multilingual text-image pairs using sigmoid loss and self-distillation methods, unlike CLIP, which employs only InfoNCE loss. 
%
However, these models rely on a large text encoder, where the size of the text encoder accounts for 64\% of the total parameters.
We require an exploration of how to integrate the small model into these techniques.

\subsection{Knowledge Distillation}

Knowledge Distillation (KD)~\cite{hinton2015distilling} is a cross-architecture knowledge transfer technique that transfers knowledge from a teacher model to a student model. 
KD methods achieve this by guiding the student's learning process to align its representations with those of the teacher, using a training objective that minimizes the discrepancy between these two models.
A common technique in the KD manner is to train the student model to maximize the similarity between the teacher and student probability distributions.  
Feature-based distillation (FD)~\cite{romero2015fitnetshintsdeepnets} aligns the teacher and student models’ representations by minimizing the mean squared error between the embeddings of the teacher and student. 
CRD~\cite{clkd} using the contrastive objective as a learning function to align teacher and student representations. 
RKD~\cite{relationalknowledgedistillation,image-rkd} aligns the teacher and student feature representations with the transfer mutual relation of the feature from the teacher to the student model.
DualL2~\cite{duall2} uses FD to minimize the mean squared error between the teacher’s English sentence representation and the student’s parallel language sentence.

To enhance the performance of knowledge distillation (KD) methods, researchers have proposed various techniques in the KD pipeline. 
For example, adding augmentation techniques to generalize students’ representation~\cite{tinybert}, improve domain-specific KD method~\cite{multilingualvqa}, or use in low-resources language KD~\cite{multilingualclkd}. Moreover, researchers employ self-distillation with a momentum encoder~\cite{ALBEF} and add an instance queue to increase the diversity of negative samples for the KD loss~\cite{seed,congen}. 
While knowledge distillation has proven effective for improving both cross-lingual and small models, the most effective approach for the multilingual vision-text encoder remains an open question.

\section{Methodology}
\label{sec:method}

\begin{figure*}
    \centering
    \includegraphics[width=\linewidth]{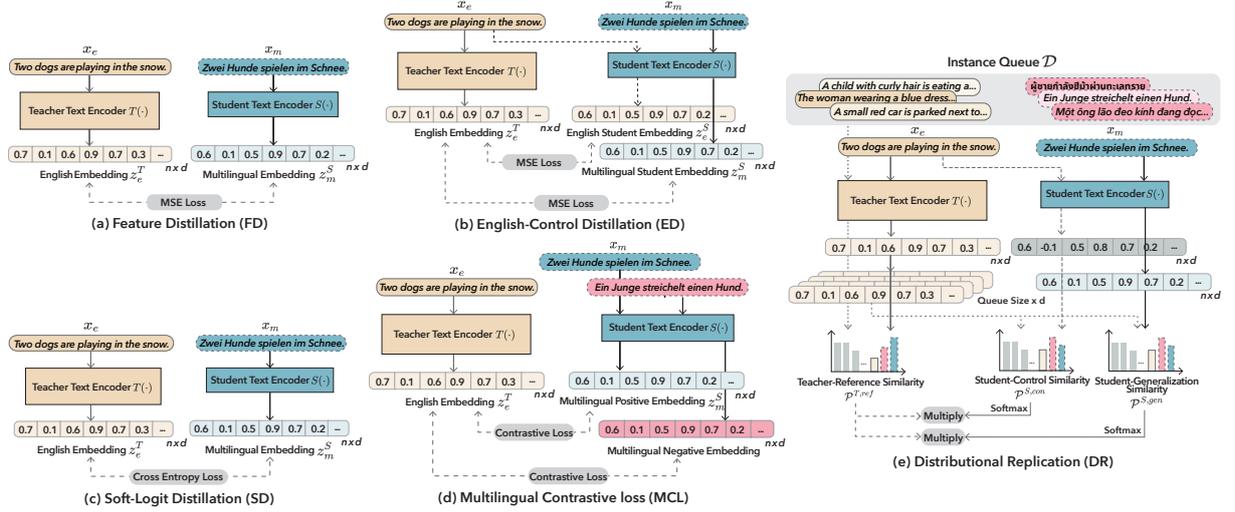}
    \vspace{-8mm}
    \caption{Illustration of variation multilingual vision-language embedding distillation in this paper}
    \vspace{-3mm}
    \label{fig1:}
\end{figure*}




\subsection{Problem Formulation}


To decrease the model's parameters, we apply the concept of knowledge distillation to transfer the knowledge from a larger model to a smaller one.
In particular, we minimize the discrepancy between large and small models, where the input can be more than one language for the student model.
%
Let $\mathcal{D} = \{(x_{i,\mathrm{e}}, x_{i,\mathrm{m}})\}_{i=1}^{N}$ denote a dataset consisting of $N$ paired English and multilingual text samples.
The vector representations produced by the student model are obtained via the embedding function $f(\cdot; \theta_\mathrm{S})$, yielding $z_{i,\mathrm{e}}^\mathrm{S} = f(x_{i,\mathrm{e}}; \theta_\mathrm{S})$ for English inputs and $z_{i,\mathrm{m}}^\mathrm{S} = f(x_{i,\mathrm{m}}; \theta_\mathrm{S})$ for multilingual inputs, where $\theta_\mathrm{S}$ denotes the parameters of the student model.
Similarly, the English representation from the teacher model, parameterized by $\theta_\mathrm{T}$, is given by $z_{i,\mathrm{e}}^\mathrm{T} = f(x_{i,\mathrm{e}}; \theta_\mathrm{T})$.

To facilitate knowledge transfer from the teacher to the student, the discrepancy between their representations is minimized using multiple objective functions, formally expressed as
$$
\min_{\theta_\mathrm{S}} \frac{1}{B} \sum_{i=1}^{B} \mathcal{L} \left(z_{i,\mathrm{m}}^\mathrm{S}, z_{i,\mathrm{e}}^\mathrm{S}, z_{i,\mathrm{e}}^\mathrm{T}; \theta_\mathrm{S}\right),
$$
where $B$ denotes the batch size.
In all experiments, the optimization objective is to minimize the loss with respect to the student model parameters $\theta_\mathrm{S}$, while keeping the teacher model parameters $\theta_\mathrm{T}$ fixed.

%



\subsection{Knowledge Distillation Loss}

As shown in Figure~\ref{fig1:}, in this study, we examine alternative loss functions for multilingual vision–language embedding distillation. 
Our investigation centers on various feature distillation approaches, contrastive learning, and distributional replication loss, with their effectiveness evaluated across multilingual benchmarks.

\subsubsection{Feature Distillation (FD)}

A straightforward approach to transferring knowledge from the teacher model to the student model is to anchor the English representations generated by the teacher and minimize their discrepancy with the student’s representations for the corresponding multilingual inputs.

The discrepancy between teacher and student representations is minimized using the Mean Squared Error loss, defined as:
\begin{equation}
    \mathcal{L}_\text{FD} = \frac{1}{B}\sum_{i=1}^{B} \left\| z_{i,\text{m}}^\text{S} - z_{i,\text{e}}^\text{T} \right\|_2^2
    \label{eq:mse_loss}
\end{equation}

\subsubsection{English-Control Distillation (ED)}

This approach extends Feature Distillation by incorporating the alignment of the student’s English text representations in addition to its multilingual representations. The inclusion of English representation alignment helps prevent representation collapse, which occurs when different language representations are drawn toward the same anchor, namely the teacher’s English representation.
Derived from feature-based distillation, the training objective is formalized as:

\begin{equation}
    \mathcal{L}_\text{ED} = \frac{1}{B} \sum_{i=1}^{B}\left( \left\|z_{i,\text{m}}^\text{S} - z_{i,\text{e}}^\text{T} \right\|_2^2 + \left\|z_{i,\text{e}}^\text{S} - z_{i,\text{e}}^\text{T} \right\|_2^2 \right)
\end{equation}

\subsubsection{Soft-Logit Distillation (SD)}


We also employ the Cross-Entropy loss, which is particularly suitable when the knowledge to be distilled is represented as probability distributions. 
This approach assumes that the teacher and student models produce vector representations that can be softened into categorical distributions:
$$
p_{i,\mathrm{e}}^\mathrm{T} = \mathrm{softmax}(z_{i,\mathrm{e}}^\mathrm{T}), \quad p_{i,\mathrm{m}}^\mathrm{S} = \mathrm{softmax}(z_{i,\mathrm{m}}^\mathrm{S})
$$
The soft-logit distillation loss measures the dissimilarity between the teacher and student probability distributions and is defined as:
\begin{equation}
    \mathcal{L}_\text{SD} = -\frac{1}{B} \sum_{i=1}^{B} p_{i,\text{e}}^\text{T} \log p_{i,\text{m}}^\text{S}
    \label{eq:CE_loss}
\end{equation}

\subsubsection{Multilingual Contrastive Learning (MCL)}

To align student and teacher representations, we employ contrastive learning (CL), which optimizes the objective by maximizing similarity between positive teacher–student pairs derived from the same English input, while contrasting them against other in-batch negative pairs. The objective is formulated as:
\begin{equation}
    \mathcal{L}^\text{MCL}_{i,\text{e}} = - \log \frac{\exp(\text{sim}(z_{i,\text{e}}^\text{S},z_{i,\text{e}}^\text{T})/\tau)}{\sum_{j=1}^B \exp(\text{sim}(z_{i,\text{e}}^\text{S},z_{j,\text{e}}^\text{T})/\tau)}
    \label{eq:p-english}
\end{equation}
where $\tau$ denotes the temperature parameter and cosine similarity is adopted as the similarity function.

In addition, to enable knowledge transfer from teacher to student, the student’s multilingual representation obtained from the same English input processed by the teacher is used to compute a parallel contrastive objective:
\begin{equation}
    \mathcal{L}^\text{MCL}_{i,\text{m}} = - \log \frac{\exp(\text{sim}(z_{i,\text{m}}^\text{S},z_{i,\text{e}}^\text{T})/\tau)}{\sum_{j=1}^B \exp(\text{sim}(z_{i,\text{m}}^\text{S},z_{j,\text{e}}^\text{T})/\tau)}
    \label{eq:p-multi}
\end{equation}

The overall Multilingual Contrastive Learning (MCL) loss combines both English-based and multilingual objectives, expressed as:
\begin{equation}
    \mathcal{L}_\text{MCL} = \frac{1}{B}\sum_{i=1}^B (\mathcal{L}^\text{MCL}_{i,\text{e}} + \mathcal{L}^\text{MCL}_{i,\text{m}}) / 2
    \label{eq:mcl} 
\end{equation}

\subsubsection{Distributional Replication (DR)}



Distributional Replication (DR) quantifies the divergence between teacher and student outputs by constructing similarity-based probability distributions. These distributions are generated from a FIFO queue of negative samples, $\bold{Q} = [q_1, ..., q_K]$, which is continuously updated with the teacher’s in-batch English representations, $[z^\text{T}_{1,\text{e}}, ..., z^\text{T}_{B,\text{e}}]$.
The generic probability distribution is defined as:
\begin{equation}
    \mathcal{P}_{ik}(\bm{z}, \bm{Q}, \tau) = \frac{\exp(\mathrm{sim}(z_i, q_k)/\tau)}{\sum_{j=1}^K \exp(\mathrm{sim}(z_i, q_j)/\tau)}
\end{equation}

Within this framework, DR specifies three distinct distributions, each serving a complementary role:
\noindent (i) Teacher-Reference distribution, computed from $\bm{z}^{\text{T}}_{e}$, which acts as the teacher-provided reference:
\begin{equation}
 \mathcal{P}^{\text{T},\text{ref}}_{ik} = \mathcal{P}_{ik}(\bm{z}^\text{T}_{\text{e}}, \bm{Q}, \tau^\text{T})
\end{equation}

\noindent (ii) Student-Control distribution, computed from $\bm{z}^{\text{S}}_{e}$, which constrains student representations to remain aligned with teacher knowledge:
\begin{equation}
 \mathcal{P}^{\text{S},\text{con}}_{ik} = \mathcal{P}_{ik}(\bm{z}^\text{S}_{\text{e}}, \bm{Q}, \tau^\text{S})
\end{equation}

\noindent (iii) Student-Generalize distribution, computed from $\bm{z}^{\text{S}}_{m}$, which facilitates broader generalization in the student’s multilingual space:
\begin{equation}
 \mathcal{P}^{\text{S},\text{gen}}_{ik} = \mathcal{P}_{ik}(\bm{z}^\text{S}_{\text{m}}, \bm{Q}, \tau^\text{S})
\end{equation}

The DR objective consists of two complementary components. The control objective enforces consistency between student-control and teacher-reference distributions:
\begin{equation}
 \mathcal{L}^\text{con}_i = - \sum_{k=1}^K \mathcal{P}^{\text{T},\text{ref}}_{ik} \log \mathcal{P}^{\text{S},\text{con}}_{ik}
\end{equation}
while the generalization objective extends this consistency to the student-generalize distribution:
\begin{equation}
 \mathcal{L}^\text{gen}_i = - \sum_{k=1}^K \mathcal{P}^{\text{T},\text{ref}}_{ik} \log \mathcal{P}^{\text{S},\text{gen}}_{ik}
\end{equation}

Finally, the overall DR loss is expressed as the average of the two objectives:
\begin{equation}
    \mathcal{L}_\text{DR} = \frac{1}{B}\sum_{i=1}^B \left(\mathcal{L}^\text{con}_i + \mathcal{L}^\text{gen}_i\right) / 2
    \label{eq:dr} 
\end{equation}


\subsection{Multi-Objective Training}

As we discussed the benefits and strengths of each training objective, we found that each loss has a trade-off, and there is no universal solution to the vision-text representation problem.
Alternatively, \cite{clip-kd, mccrolin} demonstrate the possibility of combining each training loss as a multi-task training objective.
Therefore, we summarize all training objectives with a joint knowledge distillation objective in this section:

\begin{equation}
\mathcal{L} = \sum_{i=0}^n \lambda_i \mathcal{L}_i
\end{equation}
Where \(\mathcal{L}_i\) represents the distillation objective that we mentioned previously, and \(\lambda_i\) are their weight for each objective.

\begin{table*}[h!]
    \centering
    \scalebox{0.8}{
    \begin{tabular}{l|cccccccc|ccc}
        \toprule
        & \multicolumn{8}{c|}{Retrieval} & \multicolumn{3}{c}{VQA} \\ 
        & \multicolumn{2}{c}{Multi30k} & \multicolumn{2}{c}{COCO} & WIT & xFlickr & XM3600 &  & CVQA &ALM-Bench & \\
        Methods & I2T & T2I & I2T & T2I & I2T & I2T & I2T & AVG. & Acc & Acc& AVG.\\
        \midrule
        T: SigLIP2-L/16 & 71.37 & 71.23 & 45.82 & 20.37 & 42.39 & 53.64 & 53.75 & 51.22 & 32.88 & 39.53 & 36.21\\
        \hline 
        S: XLM-R\textsubscript{Base} & & & & & & & & &\\
        +FD & \underline{69.27} & 73.20 & \textbf{36.12} & 25.89 & 21.99 & 62.24 & 48.45 & 48.17 & \underline{30.30} & 36.70 & \underline{33.50}\\
        +ED & 63.97 & \underline{74.67} & \underline{32.58} & \textbf{29.20} & \underline{30.26} & \underline{65.24} & \underline{52.65} & \underline{49.80} & \textbf{31.01} & \textbf{40.06} & \textbf{35.54}\\
        +SD & 58.80 & 68.37 & 27.12 & 24.14 & 20.53 & 57.59 & 44.62 & 43.02 & 29.49 & \underline{36.73} & 33.11 \\
        +MCL & 54.37 & 63.27 & 26.92 & 24.63 & 19.68 & 54.24 & 42.36 & 40.78 & 26.87 & 35.81 & 31.34\\
        +DR & \textbf{70.57} & \textbf{76.03} & 30.34 & 27.39 & \textbf{33.40} & \textbf{66.66} & \textbf{54.13} & \textbf{51.22} & 28.65 & 36.48 & 32.57\\
        \hline
        +DR+ED & \textbf{71.03} & 76.17 & 30.44 & 27.17 & 33.41 & 66.48 & 54.22 & 51.27 & 28.76 & 36.13& 32.45\\        
        +DR+FD & 70.47 & 76.23 & \textbf{31.08} & \textbf{27.30} & \textbf{33.64} & \textbf{66.51} & 54.23 & \textbf{51.35} & 28.67 & 36.46 & 32.57\\
        +DR+ED+FD & 70.90 & \textbf{76.53} & 30.14 & 26.51 & 33.20 & 66.22 & \textbf{54.26} & 51.11 & \textbf{28.68} & \textbf{37.53} & \textbf{33.11}\\
        \bottomrule
    \end{tabular}}
    \vspace{-2mm}
    \caption{This table presents our qualitative evaluation of two metrics: (1) retrieval Recall@1 (R@1) scores on Multi30K, COCO, WIT, xFlickr, and XM3600; and (2) multiple-choice visual question answering accuracies on CVQA and ALM-Bench. The results are from the XLM-R\textsubscript{Base} student model trained with knowledge distillation from the SigLIP2-L/16 teacher model.}
    \vspace{-3mm}
  \label{tab:1}
\end{table*}

\section{Experimental Setup}

\subsection{Training Dataset}
Following previous works~\cite{multilingualclip,mclip,siglip}, we utilize a common training dataset, Imagecaptioning7M, which comprises 7M multilingual-English text pairs. 
We used the translated version from \citet{multilingualclip}. 
This dataset was derived from sources like Google Conceptual Caption (GCC)~\cite{GCC}, MSCOCO~\cite{mscoco}, and VizWiz~\cite{vizwiz}. 
For the Validation dataset, we use a validation set of Multi30k~\cite{Multi30K}, a multilingual version of Flickr30k~\cite{flickr30k}.

\subsection{Models}
We utilize CLIP~\cite{clip}, a monolingual foundation model, and SigLIP2~\cite{siglip2}, a multilingual foundation model. 
In particular, we select CLIP-ViT-L/14 and SigLIP2-L16 as teacher models for our experiments, and we select XLM-R\textsubscript{Base}~\cite{xlmr}, MiniLM~\cite{minilm}, and DistilBERT~\cite{distillbert} as student models to mimic the teacher text's representation.
We describe the hyper-parameter settings in Appendix~\ref{appendix:setup}.

\subsection{Evaluation Benchmark} 

Similar to previous works' setting~\cite{clip,siglip,siglip2}, we evaluate our student models on seven benchmarks that cover text-image retrieval and Visual Question Answering (VQA) tasks. 
For retrieval downstream tasks, we use Multi-30k~\cite{Multi30K}, MSCOCO~\cite{mscoco}, WIT~\cite{WIT}, xFlickr~\cite{xFlickr}, and XM3600~\cite{xm3600}.
For VQA, we utilize CVQA~\cite{cvqa}, which comprises user-submitted photos and questions in 31 languages, and ALM-Bench~\cite{almbench}, which offers domain-specific cultural questions in 100 languages. 

\subsection{Evaluation metrics}

We employ Recall@k ($R@k$) for Text-to-Image (T2I) and Image-to-Text retrieval (I2T) tasks. 
Our primary metric is $R@1$, which measures top-1 accuracy, and we also report results for $R@5$ and $R@10$ in Appendix~\ref{appendix:recall}. 
For VQA, we formulate this task as a similarity matching problem similar to \citet{cvqa}. 
In particular, to determine the model answer, we concatenate the question with candidate answers, compute the cosine similarity between the combined text and image, and select the most similar choice as the answer. 
Then, we use accuracy as the main metric of VQA benchmarks. 

\section{Experimental Results}

In this section, we propose studies to explore the performance of small models using various proposed KD methods, aiming to answer \textbf{RQ1}.
In Section~\ref{subsec:main_results}, we propose an empirical study of the effectiveness and generalization of using various KD methods on the baseline model, XLM-R\textsubscript{Base}.
Section~\ref{subsec:model_variants}, we study the robustness of the optimal KD approaches through various small models.
Section~\ref{subsec:ablation}, we study the design choices of our KD method, specifically the language anchor for KD and the representation of images versus text as the anchor.
\subsection{Main Results} \label{subsec:main_results}

\begin{table*}[h!]
    \centering
    \setlength{\tabcolsep}{3pt} 
    \scalebox{0.7}{
    \begin{tabular}{l|cccccccccccc|cccccc}
        \toprule
        & \multicolumn{12}{c|}{Retrieval (I2T)} & \multicolumn{6}{c}{VQA} \\ 
        & \multicolumn{2}{c}{Multi30k} & \multicolumn{2}{c}{COCO} & \multicolumn{2}{c}{WIT} & \multicolumn{2}{c}{xFlickr} & \multicolumn{2}{c}{XM3600} &\multicolumn{2}{c|}{AVG} & \multicolumn{2}{c}{CVQA} &\multicolumn{2}{c}{ALM-Bench} &\multicolumn{2}{c}{AVG} \\
        Methods & En & Mul & En & Mul & En & Mul& En & Mul& En & Mul& En & Mul & En & Mul & En & Mul & En & Mul \\
        \midrule
        SigLIP2-L/16 & 79.60 &71.37 &70.48 &45.82 &70.70 &42.39 &74.25 &53.64 &65.72 &53.75 &72.15&53.39&31.33 &32.88& 35.88& 39.53 &33.61&36.21\\
        \hline
        \multicolumn{15}{l}{T: SigLIP2-L/16 / S: XLM-R\textsubscript{Base} (Parameters from 881M to 594M):Text Encoder from 565M to 278M }  \\
        \hline
        +FD & 75.80 & 69.27 & \textbf{63.04} & \textbf{36.12} & 35.00 & 21.99 & 71.95 & 62.24 & 49.76 & 48.45 &59.11	&47.61& \textbf{31.72} & \textbf{30.30} & 35.54 & \textbf{36.70} & \textbf{33.63} & \textbf{33.50}\\
        +DR & \textbf{76.50} & \textbf{70.57} & 63.02 & 30.34 & \textbf{51.30} & 33.40 & \textbf{75.35} & \textbf{66.66} & 57.38 & 54.13 &\textbf{64.71}	&51.02& 31.12 & 28.65 & 35.42 & 36.48 &33.27&32.57\\
        +DR+FD & 75.90 & 70.47& 62.94& 31.08& 50.70 & \textbf{33.64}& 74.90 & 66.51 & \textbf{57.50} & \textbf{54.23} &64.39 & \textbf{51.19} & 30.92& 28.68& \textbf{35.84}	&36.47 &33.38&32.58\\
        \hline 
        \multicolumn{15}{l}{T: SigLIP2-L/16 / S: DistillBert (Parameters from 881M to 450M):Text Encoder from 565M to 134M}  \\
        \hline
        +FD & 69.60 & 61.30 & 59.24 & \textbf{19.48} & 32.30 & 18.03 & 68.60 & 56.41 & 46.74 & 38.20 &55.30	&38.68& 30.39 & \textbf{28.60} & 34.37 & \textbf{37.71} & 32.38	& \textbf{33.16}\\
        +DR & 74.20& 66.27 & 61.32 & 13.64  & 45.00  & 26.11  & 74.10 & \textbf{62.66} & 56.18 & \textbf{45.30} &62.16	&42.80& \textbf{30.79} & 28.52 & \textbf{36.54} & 35.22 &\textbf{33.67}&31.87\\
        +DR+FD& \textbf{75.00} & \textbf{66.63} & \textbf{61.58} & 13.72& \textbf{45.90} & \textbf{26.76} & \textbf{74.95} & 62.53& \textbf{56.72} & 45.27&\textbf{62.83}&\textbf{42.98}&30.49 &27.54 & 34.60 & 35.34 &32.55&31.44\\
        \hline 
        \multicolumn{15}{l}{T: SigLIP2-L/16 / S: MiniLM (Parameters from 881M to 433M):Text Encoder from 565M to 117M}  \\
        \hline
        +FD & 66.80 & 58.87 & 52.68 & \textbf{29.10} & 17.30 & 11.27 & 63.75 & 50.08 & 40.22 & 37.50& 48.15	&37.36 & 27.80 & \textbf{28.52} & 30.90 & \textbf{35.81} & 29.35 & \textbf{32.17} \\
        +DR & 73.10 & 65.07 & 58.92 & 21.90  & 40.30  & 22.84 & 73.80 & 61.16 & \textbf{55.79} & \textbf{46.73} &60.38&\textbf{43.54}& 29.42 & 28.39 & \textbf{33.60} & 34.28 &\textbf{31.51}&31.34\\
        +DR+FD& \textbf{73.60} & \textbf{65.43} & \textbf{59.60} & 20.76& \textbf{41.60} & \textbf{23.08} & \textbf{74.35} & \textbf{61.56} & 55.58& 46.68&\textbf{60.95}&43.50& \textbf{29.85} &28.12& 32.87	&34.28&31.36&31.20\\
        \hline 
        CLIP-ViT-L/14 & 68.80 &-& 56.32&-& 69.50 &-&56.00 &-&42.47 &-&58.62&-&36.72 &-&43.69&-&40.21&-\\
        \hline
        \multicolumn{15}{l}{T: CLIP-ViT-L/14 / S: XLM-R\textsubscript{Base} (Parameters from 427M to 581M):Text Encoder from 123M to 278M}  \\
        \hline
        +FD & \textbf{67.00} & \textbf{62.13} & \textbf{53.40} & \textbf{33.18} & 32.00 & 22.06 & 56.80 & 47.68 & 37.50 & 36.39& 49.34&40.29 & \textbf{29.28} & \textbf{29.04} & 33.87 & 36.04  & \textbf{31.58} & \textbf{32.54}\\
        +DR & 63.30 & 59.30 & 49.04 & 25.02 & \textbf{54.60} & 34.27 & \textbf{64.40} & 53.85 & \textbf{43.72} & 42.07 &\textbf{55.01}&42.90& 28.74 & 28.13 & 33.30 & 36.34&31.02&32.24 \\
        +DR+FD& 62.70& 59.30& 49.24&25.94 &52.40 & \textbf{34.36} & 63.10& \textbf{54.01} & 43.61&\textbf{42.15}&54.21&\textbf{43.15}&28.90&27.92&\textbf{34.17}&	\textbf{36.70} &31.54&32.31 \\
        \hline 
        \multicolumn{15}{l}{T: CLIP-ViT-L/14 / S: DistillBert (Parameters from 427M to 437M):Text Encoder from 123M to 134M}  \\
        \hline
        +FD & \textbf{63.50} & 57.20 & \textbf{50.94} & \textbf{20.04} & 31.20 & 18.34 & 52.45 & 41.93 & 35.04 & 28.93& 46.63&33.29 & \textbf{29.14} & \textbf{27.75} & 32.78 & \textbf{35.54} & \textbf{30.96} & \textbf{31.65}\\
        +DR & 63.20 & 55.43 & 48.58 & 12.84 & \textbf{49.10} & 29.52 & 62.80 & \textbf{51.84} & \textbf{43.08} & 35.94 &\textbf{53.35}&37.11& 27.37 & 27.33 & \textbf{32.85} & 34.92&30.11&31.13\\
        +DR+FD& 62.20& \textbf{55.90} & 47.88& 12.28&48.80 &\textbf{30.57} & \textbf{62.90}& 51.21& 42.83&\textbf{35.96}&52.92&\textbf{37.18}&27.83&27.18&32.82	&35.24&30.33&31.21\\
        \hline 
        \multicolumn{15}{l}{T: CLIP-ViT-L/14 / S: MiniLM (Parameters from 427M to 420M):Text Encoder from 123M to 117M}  \\
        \hline
        +FD & 58.40 & 54.50 & \textbf{47.64} & \textbf{28.58} & 17.80 & 11.37 & 46.75 & 37.00 & 30.07 & 28.05 &40.13&31.90& 27.49 & \textbf{28.44} & 30.11 & \textbf{34.86} &28.80&\textbf{31.65}\\
        +DR & \textbf{62.50}  & \textbf{55.20} & 46.00 & 21.30 & \textbf{40.10} & 25.18 & 62.00 & 50.23 & 42.53 & 37.50 &\textbf{50.63}&\textbf{37.88}&\textbf{28.21} & 26.74 &31.14 & 34.24&\textbf{29.68}&30.49\\
        +DR+FD& 62.00&53.97 &46.22 & 21.28&39.40 &\textbf{25.51} & \textbf{62.40}& \textbf{50.91}& \textbf{42.63}&\textbf{37.52}&50.53&37.84&27.78&27.36&\textbf{31.39}	&34.63&29.59&31.00 \\
        \bottomrule
    \end{tabular}}
    \vspace{-2mm}
    \caption{A comparison of knowledge distillation performance for various teacher-student models trained on the ImageCaptioning7M dataset and validated on the Multi30k dataset.}
    \vspace{-3mm}
  \label{tab:2}
\end{table*}


\paragraph{KD results}
The results of our KD methods are shown in Table~\ref{tab:1}. 
We can see that the student model with the DR method outperformed other individual KD methods. 
Although the ED method performed less effectively than the DR method on retrieval tasks, it significantly outperformed all other methods on the VQA task.
For example, the ED method outperforms the DR method by 2.97 points in the VQA benchmarks.
This emphasizes that there is no universal method; the DR method is suitable for in-domain downstream tasks, such as the retrieval task, whereas ED is more generalized than DR in comparison to out-of-domain VQA tasks. 
%

\noindent
\textbf{Compare with the teacher's performance}
%
When we compare the performance of the best performing students and the teacher model, we found that \emph{the student can perform similarly to the teacher model in the average score.}
As shown in Table~\ref{tab:1}, the DR method achieves 51.22 points on the retrieval benchmarks, matching the performance of the teacher model.
We observe a reasonable improvement on all T2I experiments, including Multi30k and COCO.
%
%
These findings demonstrate that KD can enable student models to mimic teacher behavior and give better performance than the teacher in some downstream tasks. 
%

\noindent
\textbf{Combining multiple KD training objectives}
We also conduct an experiment using multi-KD training objectives in our study by combining the most effective training objectives in Table~\ref{tab:1}. 
The experimental results demonstrate that combining DR and FD yields a better improvement for the retrieval task, improving from 51.22 points with DR to 51.35 points with DR+FD.
However, we observe a performance penalty in the VQA task from 35.54 (ED) to 33.11 points (DR+ED+FD).
These findings emphasize the importance of the training objective, which is designed for the retrieval task, rather than the VQA task.
This suggests the need to develop a new training objective that effectively addresses both retrieval and VQA tasks.
Note that we demonstrate the full results of each language in Appendix~\ref{appendix:LanguageResult}.


\subsection{Model Variants} \label{subsec:model_variants}

To assess the robustness of the KD techniques, we experiment on the same benchmarks using various teacher and student models. 
In particular, we select three KD techniques: (i) FD as a strong baseline, (ii) DR as the most effective individual approach, and (iii) the optimal multi-objective configuration, DR combined with FD.
We vary the teacher–student configurations by employing SigLIP2-L/16 and CLIP-ViT-L/14 as teacher models, while considering XLM-R\textsubscript{Base}, DistilBERT, and MiniLM as student models for comparison.
%

%

\noindent
\textbf{Comparing with SigLIP2}
As shown in Table~\ref{tab:2}, we observe that when we decrease the text encoder parameters from 565M (SigLIP2-L/16) to 278M, the multilingual performance decreases from 53.39 to 51.19 points on the retrieval benchmarks, while the performance of English decreases by 7.44 points. 
However, performance decreases when the number of parameters is reduced; for example, we observed a 16.03-point decrease in FD when using MiniLM. 
Although we can mitigate this problem with our KD techniques (DR and DR+FD), a gap still remains between the teacher and student models for the retrieval benchmarks.
In contrast, we found that only a 2.71-point difference on the VQA benchmarks.
Moreover, we obtain a significantly faster inference speed, which is preferable for real-world applications.
This can be a trade-off for efficiency vs. robustness for the retrieval task.

\begin{table*}[h!]
    \centering
    \setlength{\tabcolsep}{3pt} 
    \scalebox{0.7}{
    \begin{tabular}{l|cccccccccccc|cccccc}
        \toprule
        & \multicolumn{12}{c|}{Retrieval (I2T)} & \multicolumn{6}{c}{VQA} \\ 
        & \multicolumn{2}{c}{Multi30k} & \multicolumn{2}{c}{COCO} & \multicolumn{2}{c}{WIT} & \multicolumn{2}{c}{xFlickr} & \multicolumn{2}{c}{XM3600} & \multicolumn{2}{c|}{AVG} & \multicolumn{2}{c}{CVQA} &\multicolumn{2}{c}{ALM-Bench} & \multicolumn{2}{c}{AVG} \\
        Methods & En & Mul & En & Mul& En & Mul& En & Mul& En & Mul & En & Mul & En & Mul & En & Mul & En & Mul \\
        \midrule
        \multicolumn{18}{l}{Method: DR / T: SigLIP2-L/16 / S: XLM-R\textsubscript{Base} } \\
        \midrule
        English & \textbf{76.50} & \textbf{70.57} & 63.02 & 30.34 & \textbf{51.30} & \textbf{33.40} & \textbf{75.35} & \textbf{66.66} & \textbf{57.38} & \textbf{54.13} & \textbf{64.71} & \textbf{51.02} & \textbf{31.12} & 28.65 & \textbf{35.42} & \textbf{36.48} & \textbf{33.27} & \textbf{32.57} \\
        German & 71.40 & 70.17 & \textbf{64.54} & \textbf{37.26} & 45.70 & 30.20 & 71.10 & 64.19 &52.04 & 52.43 & 60.96 & 50.85 & 30.87 & 28.58 & 34.86 & 36.02 & 32.87 & 32.30 \\
        China & 53.90 & 47.30  & 48.12 &25.70 &25.90 & 17.35 &49.45 & 40.74 &36.76 & 36.01 & 42.83 & 33.42 & 30.55 & \textbf{28.96} &34.15 & 35.11 & 32.35 & 32.04 \\
        \midrule
        \multicolumn{18}{l}{Method: DR+FD / T: SigLIP2-L/16 / S: XLM-R\textsubscript{Base}} \\
        \midrule
        English & \textbf{75.90} & \textbf{70.47} & 62.94& 31.08& \textbf{50.70} & \textbf{33.64} & \textbf{74.90} & \textbf{66.51} & \textbf{57.50} & \textbf{54.23} &\textbf{64.39} & \textbf{51.19} & \textbf{30.92}& 28.68& \textbf{35.84} &36.47 &\textbf{33.38}&32.58\\
        German & 70.60 & 70.13 & \textbf{64.20} & \textbf{36.06} & 45.70 & 29.59 & 71.15 & 63.54 & 52.40 & 52.36 & 60.81 & 50.37 & 30.78 & \textbf{28.96} & 34.24 & \textbf{36.69} & 32.51 & \textbf{32.83} \\
        China & 53.50 & 47.90  & 48.12 &25.58 &25.70 & 17.21 & 49.55 & 41.77 & 37.12 & 35.95 & 42.80 & 33.68 & 29.90 & 28.18 & 35.26 & 35.79 & 32.58 & 31.99 \\
        \bottomrule
    \end{tabular}}
    \vspace{-2mm}
    \caption{A comparison of knowledge distillation performance in SigLIP2-L/16 as teacher and XLM-R\textsubscript{Base} as student when changing language anchor.}
    \vspace{-3mm}
  \label{tab:3}
\end{table*}

\paragraph{Cross-lingual transfer capability}
Interestingly, we found that when we use CLIP-ViT-L/14 as the teacher model (which only supports English), we can create a student model that supports multiple languages.
This is because our learning techniques did not rely on multilingual representation, but instead used only a monolingual representation, enabling the student model to learn any languages supported by the training dataset.
The experimental results demonstrate a comparable result between CLIP-ViT-L/14 and students on the English results for the retrieval benchmark.

\paragraph{Multi-training objective is essential}
When focusing on the multi-objective learning results, it provides an improvement in the setting of smaller student models and in-domain downstream tasks. 
From Table~\ref{tab:2}, multi-objective models mostly maintain English performance over single-objective approaches, specifically across all SigLIP2-DistillBert retrieval benchmarks and in four out of five SigLIP2-MiniLM retrieval benchmarks.
For CLIP-ViT-L/14 as the teacher model, the multilingual results of the multi-objective approach outperform three of five in retrieval benchmarks (WIT, xFlickr, and XM3600).
These results demonstrate that in a small student model, multi-objective learning plays a crucial role in enhancing knowledge transfer from teacher to student.
%

\subsection{Ablation study} \label{subsec:ablation}

To confirm the knowledge distillation setting in our work, we conduct ablation studies to observe the performance improvement of each component.
We observe that a major component of our KD techniques is the representation of the anchor, while we use English text as the representation for the student model to mimic.
Emergent questions raised are: (i) Can we use images instead of text representations? (ii) Is the KD framework generalized to other languages as the anchor?

\subsubsection{Non-english language anchor}

\paragraph{Setup}
While a multilingual teacher, SigLIP2-L/16, is able to encode non-English languages, we translated the training dataset into non-English languages to study the knowledge transfer performance when it comes from non-English texts. 
We select two non-English languages, German, which represents the Indo-European language family, and Chinese, which represents the Sino-Tibetan language family, as training anchors. 
Then, we translated Imagecaptioning7M to selected languages with Qwen3-4B-Instruct~\cite{qwen3technicalreport}. 
%

\begin{figure*}[h!]
    \centering
    \includegraphics[width=0.9\linewidth]{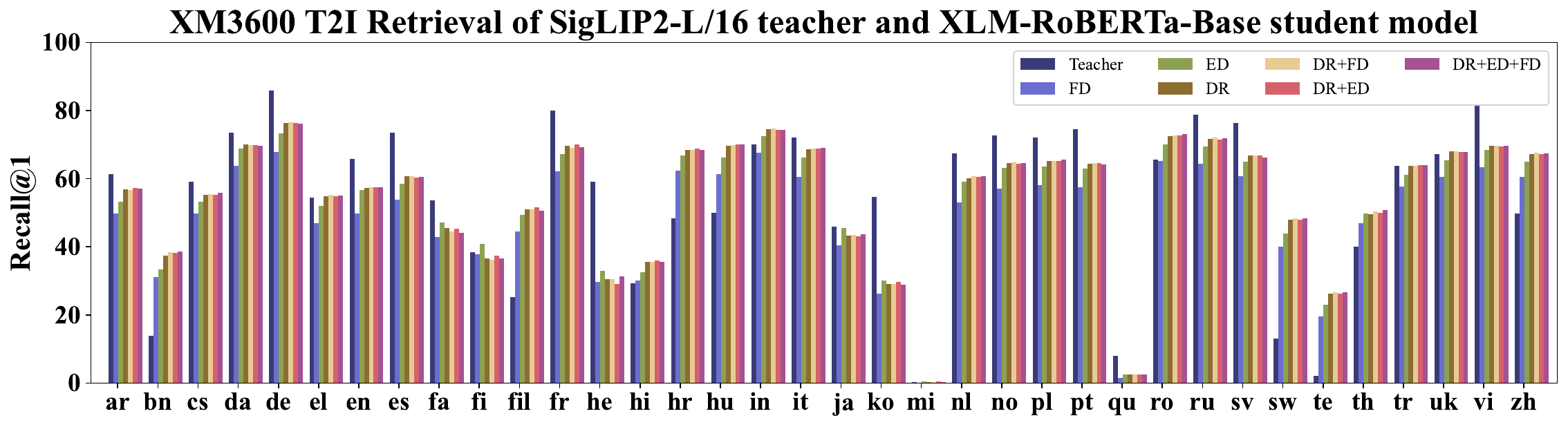}
    \vspace{-5mm}
    \caption{Result of recall@1 on the XM3600 dataset}
    \vspace{-3mm}
    \label{fig:multilingual}
\end{figure*}

\paragraph{Results}
As shown in Table~\ref{tab:3}, using non-English languages leads to performance drops compared to English. The German-trained model outperforms the Chinese-trained model by 17.42 points on the retrieval benchmark, using SigLIP2-L/16 as the teacher, although both remain below the English-trained model. 
We attribute this to linguistic proximity (German), which is related to English, preserving more knowledge during translation, while Chinese’s distinct structure causes greater information loss, as seen in Figure~\ref{fig:multilingual}.
Interestingly, VQA results show only slight declines from English, with mixed outcomes between German- and Chinese-trained models. 
Notably, in DR+FD, the German-trained model even surpasses the English model on CVQA and ALM-Bench multilingual tasks, indicating that non-English models can achieve VQA performance comparable to that of the English model and the teacher model.

\begin{table*}[h!]
    \centering
    \setlength{\tabcolsep}{3pt} 
    \scalebox{0.7}{
    \begin{tabular}{l|cccccccccccc|cccccc}
        \toprule
        & \multicolumn{12}{c|}{Retrieval (I2T)} & \multicolumn{6}{c}{VQA} \\ 
        & \multicolumn{2}{c}{Multi30k} & \multicolumn{2}{c}{COCO} & \multicolumn{2}{c}{WIT} & \multicolumn{2}{c}{xFlickr} & \multicolumn{2}{c}{XM3600} & \multicolumn{2}{c|}{AVG} & \multicolumn{2}{c}{CVQA} &\multicolumn{2}{c}{ALM-Bench} & \multicolumn{2}{c}{AVG} \\
        Anchor & En & Mul & En & Mul& En & Mul& En & Mul& En & Mul & En & Mul & En & Mul & En & Mul & En & Mul\\
        \midrule
        \multicolumn{18}{l}{Method: DR / T: SigLIP2-L/16 / S: DistillBert } \\
        \midrule
        Text & \textbf{73.80} & \textbf{63.80} & \textbf{59.04} & \textbf{14.50} & \textbf{49.10} & \textbf{25.99} & \textbf{75.00} & \textbf{60.64} & \textbf{57.65} & \textbf{45.80} & \textbf{62.92} & \textbf{42.15} & 31.12 & 28.02 & 35.77 & \textbf{36.31} & 33.46 & \textbf{32.17} \\
        Image   &49.90 &38.40 &44.36 &8.30 &30.90 &17.25 &57.60 & 42.57 &42.53 & 30.28 & 45.06 & 27.36 & \textbf{31.65} & \textbf{28.06} & \textbf{37.07} & 35.04 & \textbf{34.36} & 31.55 \\
        \midrule
        \multicolumn{18}{l}{Method: DR+FD / T: SigLIP2-L/16 / S: DistillBert } \\
        \midrule
        Text & \textbf{74.50} & \textbf{63.90} & \textbf{59.26} & \textbf{13.40} & \textbf{48.70} & \textbf{26.05} & \textbf{75.00} & \textbf{61.59} & \textbf{57.47} & \textbf{45.84} & \textbf{62.99} & \textbf{42.17} & 30.35 & 27.67 & 36.52 & 35.65 & 33.44 &31.66\\
        Image   &48.90 &37.27 &43.90 &8.34 &30.70 &17.74 &58.45 & 43.74 & 42.36 & 30.29 & 44.86 & 27.48 & \textbf{30.93} & \textbf{29.16} & \textbf{36.71} & \textbf{35.84} & \textbf{33.82} & \textbf{32.50} \\
        \bottomrule
    \end{tabular}}
    \vspace{-2mm}
    \caption{A comparison of knowledge distillation performance in SigLIP2-L/16 as teacher and DistillBert as student when using images as anchor.}
    \vspace{-2mm}
  \label{tab:4}
\end{table*}

\subsubsection{Using image representation as anchor}

\paragraph{Setup}
Since we use the image-text encoder as the teacher model, we raise the question of whether we can replace a text with an image sample from ImageCaptioning7M to improve performance.
From the datasets, we are unable to collect all images in ImageCaptioning7M from the web source, so we can collect approximately 55\% of the image data. 
Therefore, we will compare the text anchor and image anchor with the same total number of training data for a fair comparison. 

\paragraph{Results}
The results in Table~\ref{tab:4} show that the student model using an image anchor performs worse than the model using a text anchor in the retrieval benchmarks.
We observed a 14.79 point gap between text and image performance using DR, where the decreasing trend is also similar for DR+FD.
We hypothesize that the significant performance drop in retrieval tasks is due to using the image representation as an anchor. 
While the text anchor encodes semantic and grammatical information directly related to the original text, the image anchor provides more ambiguous reference information because its representations encode visual features that can correspond to multiple descriptions. 
Therefore, the model distilled with text anchors can accurately retrieve the texts or images corresponding to a given pair. 
However, in the VQA benchmark, the performance of the image-anchor and text-anchor students is comparable. 
This might be because VQA is the out-of-domain task; using text or image representations cannot mitigate this problem.

\section{Analysis} \label{sec:analysis}

\begin{figure*}[h!]
    \centering
    \includegraphics[width=0.9\linewidth]{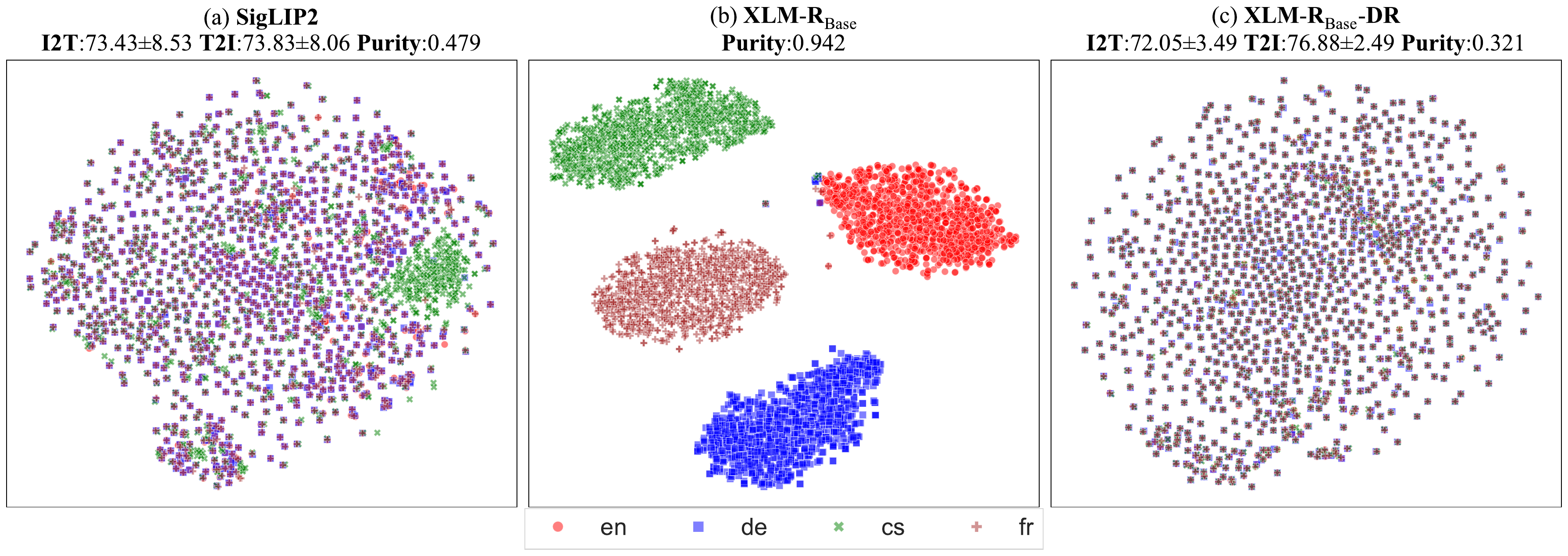}
    \vspace{-5mm}
    \caption{Embedding distribution in Multi30k dataset.}
    \vspace{-3mm}
    \label{fig:tsne}
\end{figure*}

\begin{figure*}[h!]
    \centering
    \includegraphics[width=0.9\linewidth]{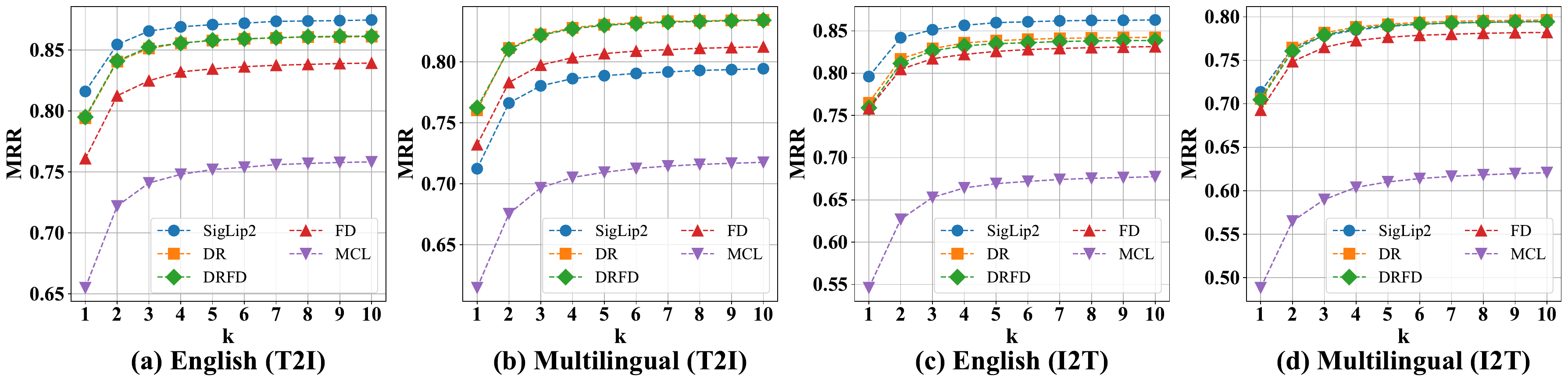}
    \vspace{-5mm}
    \caption{Result of MRR@K from Multi30k dataset.}
    \vspace{-3mm}
    \label{fig:MRR}
\end{figure*}

To better understand the results in downstream performance, we provide analyses centered around \textbf{RQ2}, as follows: 
\begin{compactitem}
    \item Section~\ref{subsec:imp_analysis}, we investigate teacher-student efficiency along with various KD techniques across various languages.
    
    \item Section~\ref{subsec:imp_analysis_xx}, we study the language distribution in latent representation.

    \item Section~\ref{subsec:ranking_analysis}, we present the analysis of ranking performance in a retrieval dataset.
\end{compactitem}

\subsection{Improvement Across Languages} \label{subsec:imp_analysis}

As shown in Table~\ref {tab:1}, we observe cases where the student model outperforms the teacher model.
This raises the question of why the student model, which has fewer parameters than the teacher model, can outperform it.
As shown in Figure~\ref{fig:multilingual}, we found that the teacher model performs poorly in certain languages.
For example, the performance of Japanese and Chinese on xFlickr is lower than that of other languages, resulting in a reasonable improvement for the student.
Additionally, the teacher model's performance is poor on Bengali, Telugu, and Swahili; however, these languages are included in our training data, which improves the performance of the student model.
This emphasizes the importance of our KD approach on small models, which, although some languages are not well-aligned, can be improved using the available languages in the training data.
%
%

\subsection{Representation Analysis} 
\label{subsec:imp_analysis_xx}

To further understand the language distribution in the latent space of a student model, we visualize its embedding using t-SNE on the Multi30k dataset, based on XLM-R\textsubscript{Base} as a student model, with the most effective approach being the DR.
Assuming that \emph{the ideal representation would be distributed well across languages}, quantified by the purity score~\cite{zhao2001criterion}, which indicates the discrepancy of its clustering performance (lower is better).

As shown in Figure~\ref{fig:tsne}a, although the retrieval performance of the teacher model is higher than the student model (Figure~\ref{fig:tsne}c), when we plot the sample using t-SNE, we observe that the clustering result of the teacher model is poorer than the student model.
We can see that the purity score of the student model is lower than that of the teacher model (0.479 vs. 0.321 points).
Although we started from a score of 0.942 (Figure~\ref{fig:tsne}b), we can enhance it to outperform the teacher model.
This emphasizes that the KD method, which focuses on improving multilingual consistency (i.e., all languages exhibit the same distribution), can yield a significant improvement in clustering results. 
%
%
We provide results from other KD methods in Appendix~\ref{appendix:embedding}.

\subsection{Ranking Robustness}
\label{subsec:ranking_analysis}

To empirically examine the robustness and consistency of these properties in downstream tasks, we analyze them via a ranking-based evaluation. 
As illustrated in Figure~\ref{fig:MRR}, the Mean Reciprocal Rank (MRR) metric is employed to quantify the performance of student models across multiple candidate retrieval tasks.
The MRR curve shows that the DR-distilled model achieves higher MRR values than other student models across all k.
%
%
Although the retrieval performance of the DR-distilled student model does not surpass that of the teacher model on the English results, this student model outperforms the SigLIP2-L/16 on the multilingual text-to-image retrieval task and yields comparable retrieval performance on the multilingual image-to-text task. 
This improvement can be attributed to the use of text-anchor embeddings during distillation, which enhances the text encoder’s capability in multilingual settings. Similar to the previous experiment, which utilized a pre-trained language model, this approach further improves performance beyond that of the original text encoder in the teacher model. 
Note that we presented the other language results in Appendix~\ref{appendix:mrr}.
%

\section{Conclusion}
We present a comprehensive study of the knowledge distillation technique in multilingual visual-language model settings. 
Our study presents the results of design choices that facilitate knowledge transfer to a small model.
The experimental results demonstrate that we can decrease the model size from 881M to 433M, where the consistency of the multilingual model is decreased by a margin on retrieval datasets, but it performs similarly on the VQA task.
%
%
We also present an analysis of performance in the student model and found that, although small models perform lower than the teacher model on retrieval and VQA benchmarks, on tasks that require multilingual consistency (e.g., clustering and ranking), student models can outperform teacher models on these tasks.

\subsection*{Acknowledgement}

This research is supported by the National Research Foundation, Singapore, under its National Large Language Models Funding Initiative. Any opinions, findings, and conclusions or recommendations expressed in this material are those of the author(s) and do not reflect the views of the National Research Foundation, Singapore.

\bibliography{tacl2021}
\bibliographystyle{acl_natbib}


\clearpage
\appendix
\onecolumn
\section{Experiment setting detail} \label{appendix:setup}
\noindent
\textbf{Model settings}
 Table~\ref{tab:setting}  reports our experiment settings in knowledge distillation methods, and Table~\ref{tab:settingDR} reports DR parameters.
All parameters are evaluated on the retrieval and VQA benchmarks.

\noindent
\textbf{Compute settings} All models were trained on a single A100 for $\sim$24-48 hrs, with the training time depending on the training method.

\noindent
\textbf{Training Language}
 The following languages are included in the ImageCaptioning 7M dataset (Training dataset): Afrikaans, Albanian, Amharic, Arabic, Azerbaijani, Bengali, Bosnian, Bulgarian, Catalan, Chinese (Simplified and Traditional), Croatian, Czech, Danish, Dutch, English, Estonian, French, German, Greek, Hindi, Hungarian, Icelandic, Indonesian, Italian, Japanese, Macedonian, Malayalam, Marathi, Polish, Portuguese, Romanian, Russian, Serbian, Slovenian, Spanish, Swahili, Swedish, Tagalog, Telugu, Turkish, Turkmen, Ukrainian, Urdu, Uyghur, Uzbek, and Vietnamese.

\begin{table}[h!]
\centering
\scalebox{0.7}{
\begin{tabular}{l|r|r|r|r}
\hline
\multicolumn{1}{c|}{Methods} & \multicolumn{1}{c|}{lr} & \multicolumn{1}{c|}{epochs} & \multicolumn{1}{c|}{batch size} & \multicolumn{1}{c}{warm up steps} \\ \hline
FD                            & \multicolumn{1}{r|}{$1e^{-5}$} & 3                     & 64                              & 1000                               \\ 
ED                            & \multicolumn{1}{r|}{$1e^{-5}$} & 3                     & 64                              & 1000                               \\ 
SD                            & \multicolumn{1}{r|}{$1e^{-5}$} & 3                     & 64                              & 1000                               \\ 
MCL                           & \multicolumn{1}{r|}{$1e^{-5}$} & 2                     & 64                              & 1000                               \\ 
DR                            & \multicolumn{1}{r|}{$1e^{-4}$} & 10 & 64                              & 1000                               \\ 
DR+FD                          & \multicolumn{1}{r|}{$1e^{-4}$} & 10 & 64                              & 1000                               \\ \hline
\end{tabular}}
\caption{Experiment settings in each KD method.}
\vspace{-3mm}
\label{tab:setting}
\end{table}

\begin{table}[h!]
\centering
\scalebox{0.7}{

\begin{tabular}{l|l|c|r|r|c|c}
\hline
\multicolumn{1}{c|}{Teacher Model} & \multicolumn{1}{c|}{Student Model} & \multicolumn{1}{c|}{Loss} & \multicolumn{1}{c|}{$\tau^{T}$} & \multicolumn{1}{c|}{$\tau^S$} & \multicolumn{1}{c|}{$K$} & \multicolumn{1}{c}{lr} \\ 
\hline
CLIP-ViT-L/14 & DistillBert & DR & 0.05 & 0.07 & 65536 & $1e^{-4}$ \\ 
CLIP-ViT-L/14 & MiniLM & DR & 0.05 & 0.07 & 65536 & $1e^{-4}$\\ 
CLIP-ViT-L/14 & XLM-R\textsubscript{Base} & DR & 0.05 & 0.07 & 65536 & $1e^{-4}$\\ 
SigLIP2-L/16 & DistillBert & DR & 0.05 & 0.07 & 65536 & $3e^{-4}$\\ 
SigLIP2-L/16 & MiniLM & DR & 0.05 & 0.07 & 65536 & $3e^{-4}$\\ 
SigLIP2-L/16 & XLM-R\textsubscript{Base} & DR & 0.05 & 0.07 & 65536 & $1e^{-4}$\\ \hline
\end{tabular}}
\caption{Distillation Hyperparameters for the DR method.}
\vspace{-3mm}
\label{tab:settingDR}
\end{table}

\section{Recall@k results}
\label{appendix:recall}
Table~\ref{tab:appendix1} reports $R@5$ and $R@10$ of retrieval benchmarks in various knowledge distillation techniques with SigLIP2-L/16 as a teacher model and XLM-R\textsubscript{Base} as a student model.
The experimental results demonstrate the consistency between $R@1$ (Table~\ref{tab:1}), $R@5$, and $R@10$, where student models perform similarly to the teacher model on the average score.

\begin{table*}[h!]
    \centering
    \scalebox{0.7}{
    \begin{tabular}{l|cccccccc}
        \toprule
        & \multicolumn{8}{c}{Retrieval (R@5)} \\ 
        & \multicolumn{2}{c}{Multi30k} & \multicolumn{2}{c}{COCO} & WIT & xFlickr & XM3600 &  \\
        Methods & I2T & T2I & I2T & T2I & I2T & I2T & I2T & AVG.\\
        \midrule
        T: SigLIP2-L/16 & 90.40  & 89.80 & 71.68 & 42.63 & 60.83 & 76.79 & 72.86 & 72.14 \\
        \hline 
        S: XLM-R\textsubscript{Base} & & & & & & \\
        +FD & 90.37 & 91.80 & 58.04 & 47.82 & 41.96 & 83.64 & 70.61 & 69.18  \\
        +ED & 89.73 & 93.13 & 57.76 & 53.08 & 52.04 & 85.28 & 74.03 &  72.15 \\
        +SD & 85.77 & 89.83 & 51.30 & 46.41 & 41.55 & 80.99 & 68.08 &  66.28 \\
        +MCL & 83.47  & 86.30 & 50.24 & 46.34 & 39.70  & 78.72 & 66.62 &   \\
        +DR & 91.73 & 93.07 & 56.28 & 49.43 & 54.03 & 85.84 & 73.64 & 72.00  \\
        \hline
        +DR+FD & 91.83 & 92.83 & 57.42 & 49.60 & 54.61 & 85.73 & 74.67 &  72.38 \\
        +DR+ED & 92.27 & 93.30 & 57.48 & 49.68 & 54.39 & 86.00 & 74.66 &  72.54  \\
        +DR+ED+FD & 91.90 & 93.10 & 54.64 & 48.80 & 53.41 & 85.33 & 74.83 &  71.72 \\
        \midrule
        & \multicolumn{8}{c}{Retrieval (R@10)} \\ 
        \midrule
        T: SigLIP2-L/16 & 94.63 & 94.10 & 80.10 & 54.01 & 68.21 & 83.97 & 78.28 & 79.04 \\
        \hline 
        S: XLM-R\textsubscript{Base} & & & & & & \\
        +FD & 94.37 & 95.83 & 67.80 & 57.78 & 52.62 & 89.31 & 77.51 &  76.46 \\
        +ED & 93.83 & 96.27 & 69.54 & 62.70 & 62.25 & 90.60 & 80.15 &  79.33 \\
        +SD & 92.13 & 94.47 & 61.62 & 56.82 & 52.60 & 87.27 & 75.55 &  74.35 \\
        +MCL & 90.60 & 92.03 & 61.68 & 56.35 & 49.81 & 86.05 & 74.79 &   \\
        +DR & 95.23 & 95.80 & 67.02 & 59.46 & 62.50 & 90.65 & 80.48 &  78.73 \\
        \hline
        +DR+FD & 95.03 & 95.87 & 67.34 & 59.20 & 63.08 & 90.86 & 80.55 &  78.85 \\
        +DR+ED & 95.27 & 96.10 & 68.22 & 59.72 & 62.94 & 90.81 & 80.52 &  79.08  \\
        +DR+ED+FD & 94.93 & 95.80 & 66.70 & 58.61 & 62.58 & 90.40 & 80.67 &  78.53 \\
        \bottomrule
    \end{tabular}}
    \caption{Retrieval result (R@5 and R@10) scores on Multi30K, COCO, WIT, xFlickr, and XM3600. The results are from the XLM-RoBERTa base student model trained with knowledge distillation from the SigLIP2-L/16 teacher model.}
    \vspace{-3mm}
  \label{tab:appendix1}
\end{table*}

\section{Embedding distribution} \label{appendix:embedding}
\begin{figure*}[h!]
    \centering
    \includegraphics[width=\linewidth]{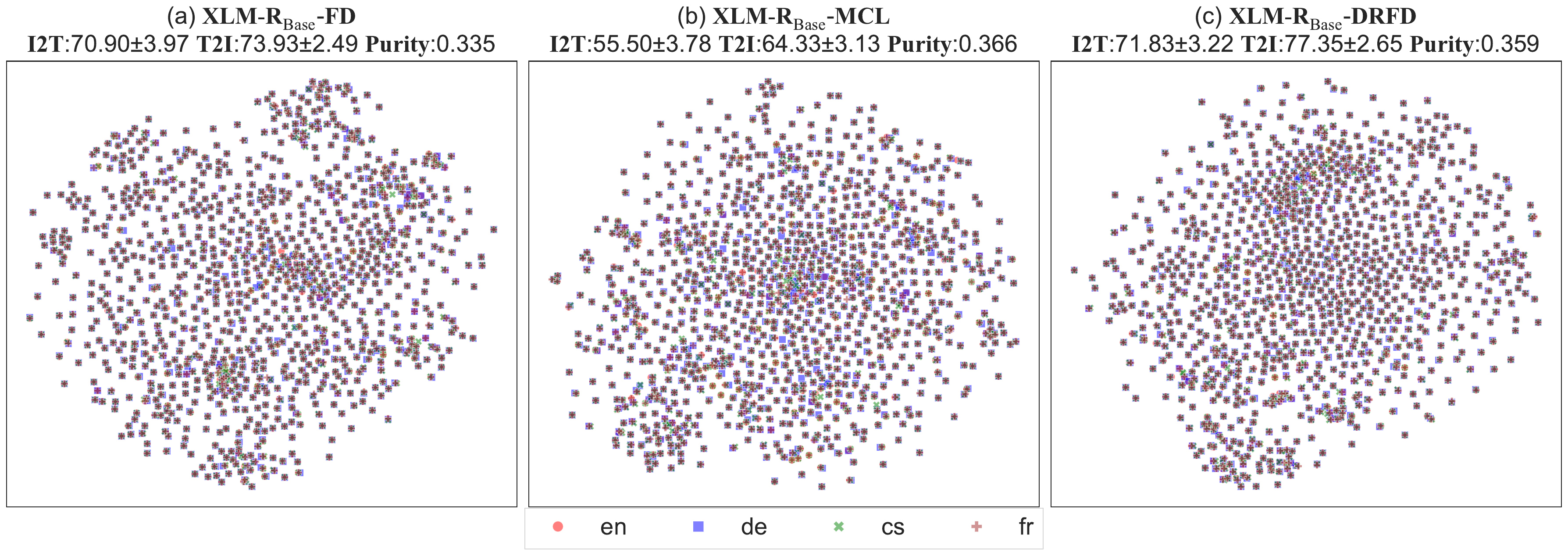}
    \vspace{-4mm}
    \caption{Embedding distribution of (a) Feature distillation (FD), (b) Multilingual Contrastive Learning (MCL), and in the Multi30k dataset.}
    \vspace{-3mm}
    \label{fig:tsne2}
\end{figure*}

Fig~\ref{fig:tsne2} reports the embedding distribution of other knowledge distillation methods.
As expected, the results of KD models are consistent in that they can perform clustering better than the teacher model, although the retrieval performance is lower than that of the teacher model.

\section{MRR@k results} \label{appendix:mrr}
\begin{figure*}[h!]
    \centering
    \includegraphics[width=\linewidth]{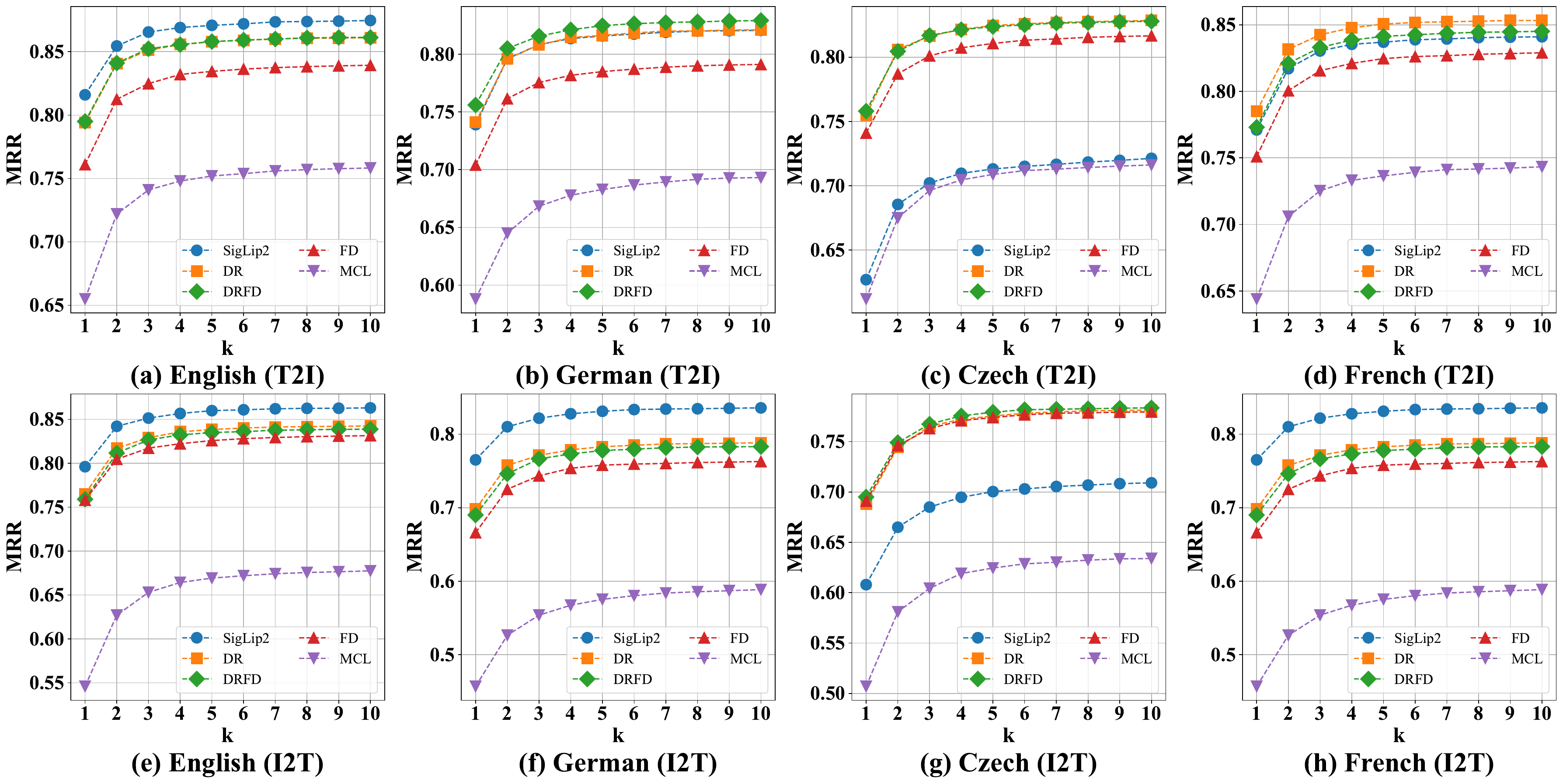}
    \vspace{-8mm}
    \caption{Result of MRR@K from Multi30k dataset.}
    \vspace{-3mm}
    \label{fig:MRR_all_language}
\end{figure*}

Fig~\ref{fig:MRR_all_language} reports specific $MRR@K$ in each language of the Multi30k dataset.
The MRR is an alternative and confirmation result for the ranking performance (Appendix~\ref{appendix:recall}).
We found that for non-English results, our KD model can perform similar to the teacher model, although the size of text encoder is reduced by half. 

\section{Language performance results} \label{appendix:LanguageResult}

\begin{figure*}[h!]
    \centering
    \includegraphics[width=\linewidth]{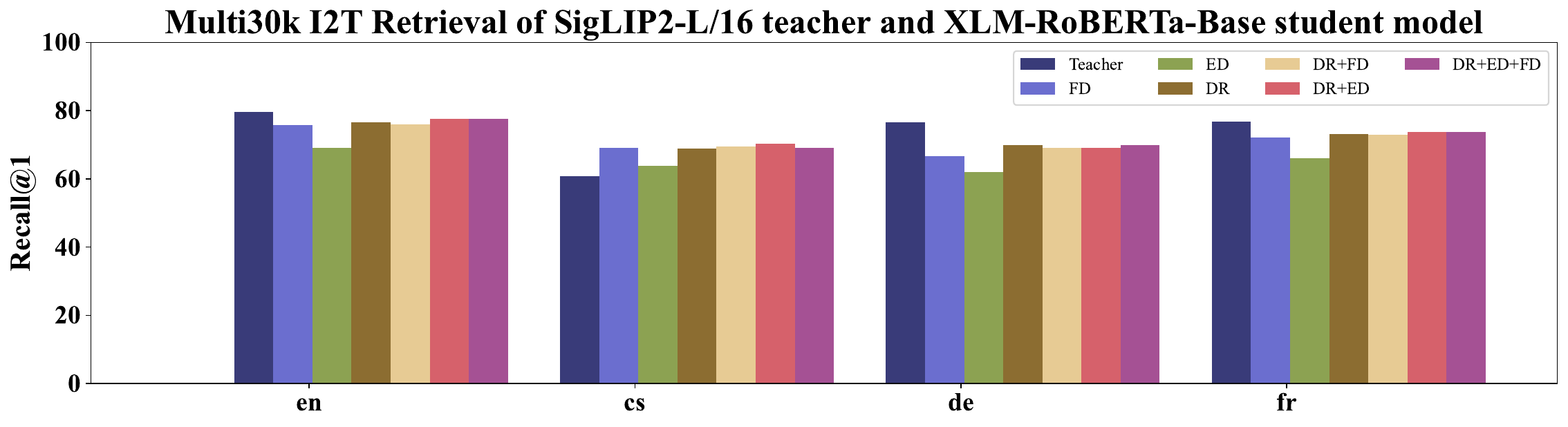}
    \vspace{-8mm}
    \caption{Result of recall@1 I2T on the Multi30k dataset.}
    \vspace{-3mm}
    \label{fig:Multi30kI2T_Time}
\end{figure*}

\begin{figure*}[h!]
    \centering
    \includegraphics[width=\linewidth]{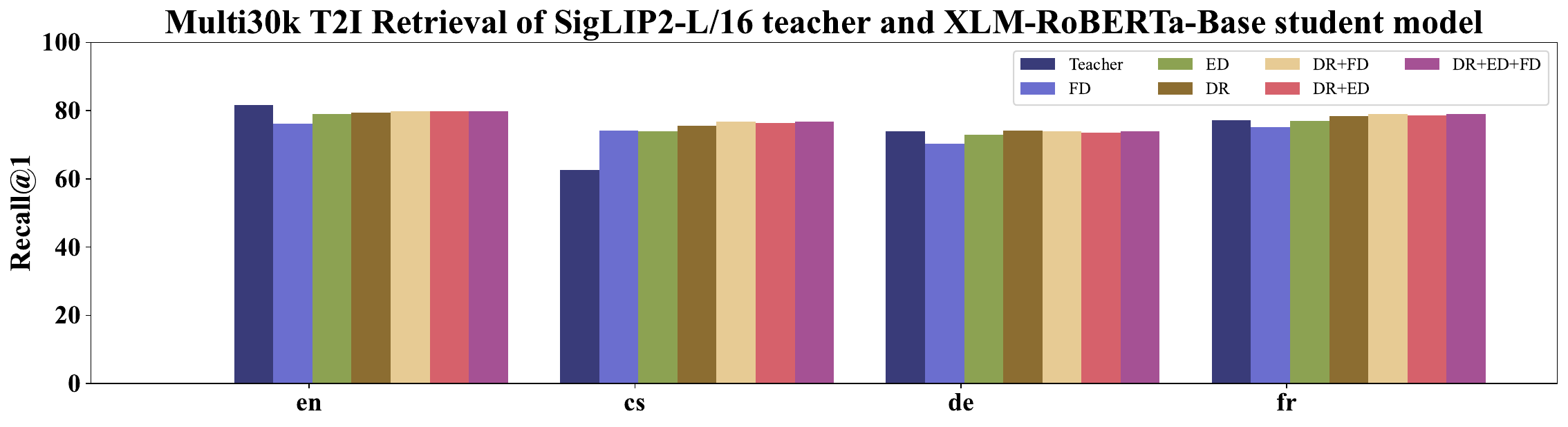}
    \vspace{-8mm}
    \caption{Result of recall@1 T2I on the Multi30k dataset.}
    \vspace{-3mm}
    \label{fig:Multi30kT2I_Time}
\end{figure*}

\begin{figure*}[h!]
    \centering
    \includegraphics[width=\linewidth]{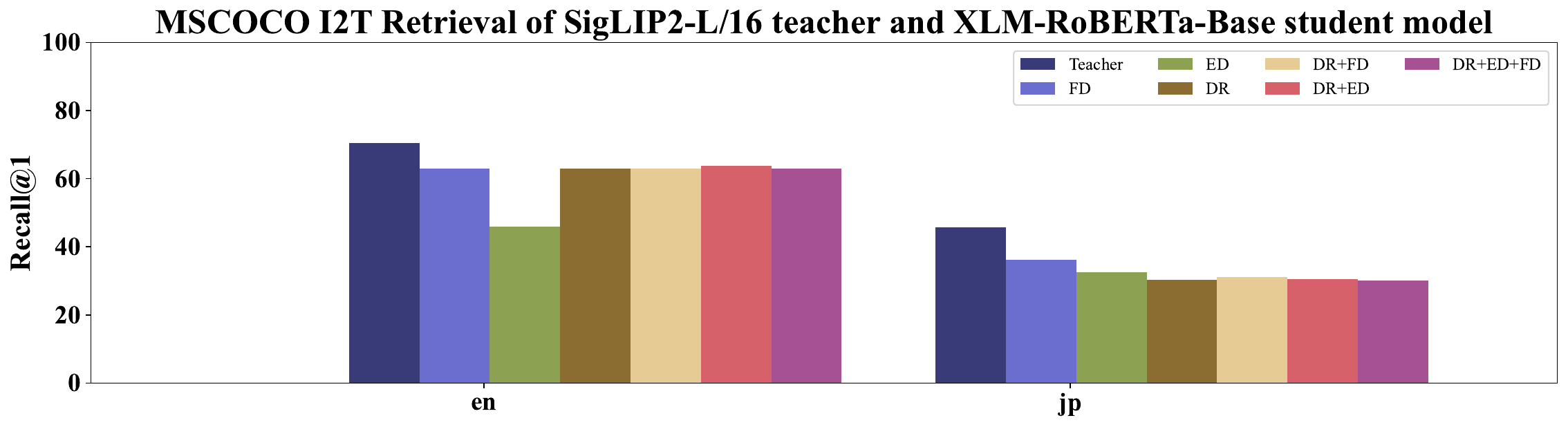}
    \vspace{-8mm}
    \caption{Result of recall@1 I2T on the MSCOCO dataset.}
    \vspace{-3mm}
    \label{fig:MSCOCOI2T_Time}
\end{figure*}

\begin{figure*}[h!]
    \centering
    \includegraphics[width=\linewidth]{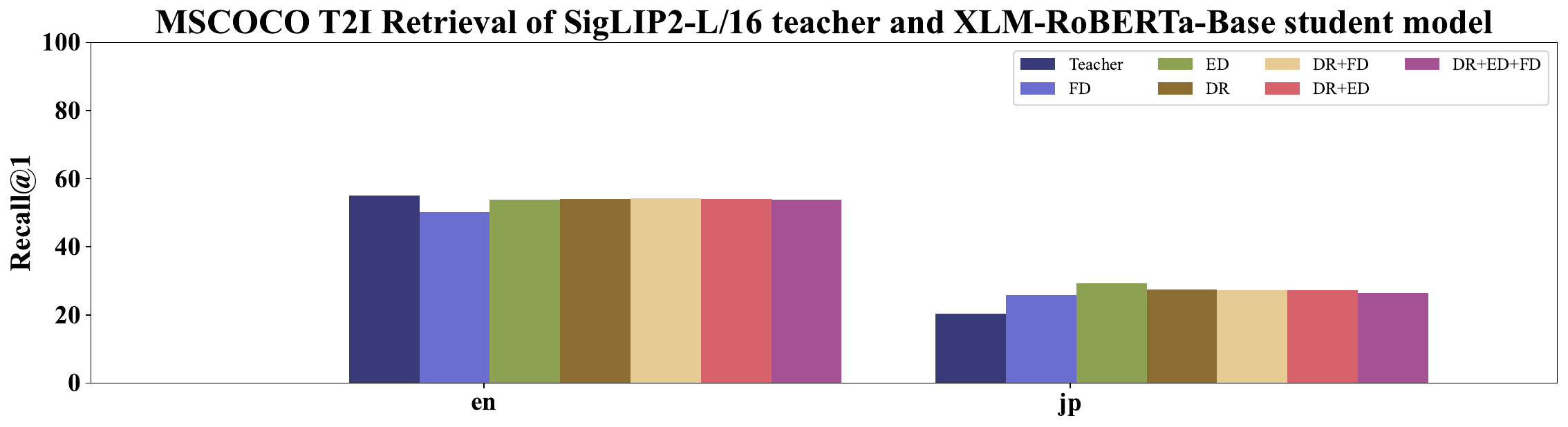}
    \vspace{-8mm}
    \caption{Result of recall@1 T2I on the MSCOCO dataset.}
    \vspace{-3mm}
    \label{fig:MSCOCOT2I_Time}
\end{figure*}

\begin{figure*}[h!]
    \centering
    \includegraphics[width=\linewidth]{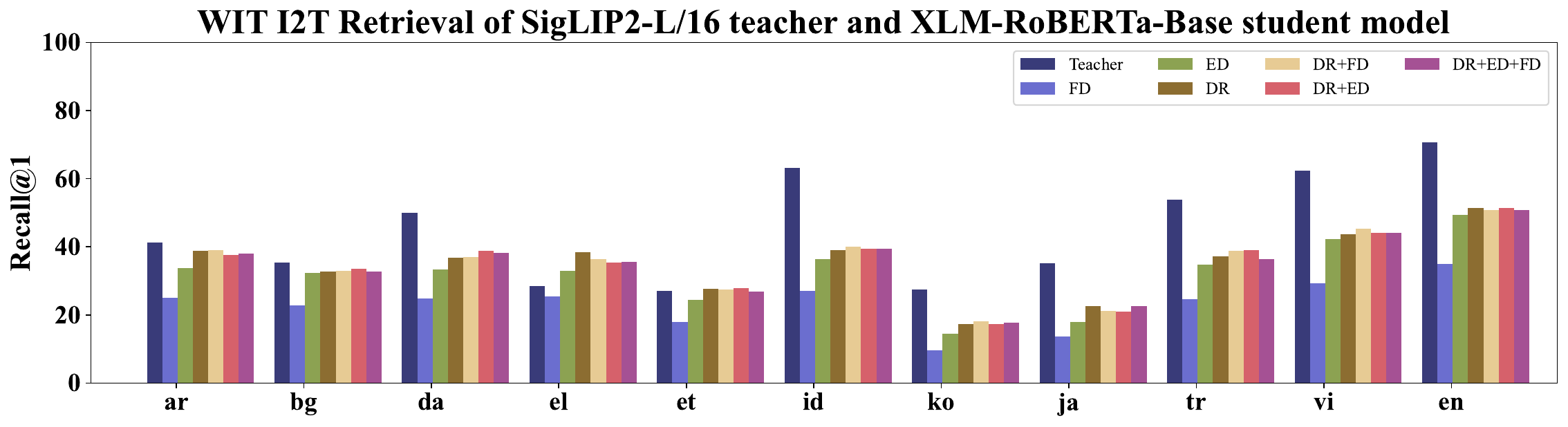}
    \vspace{-8mm}
    \caption{Result of recall@1 on the WIT dataset.}
    \vspace{-3mm}
    \label{fig:WITTime}
\end{figure*}

\begin{figure*}[h!]
    \centering
    \includegraphics[width=\linewidth]{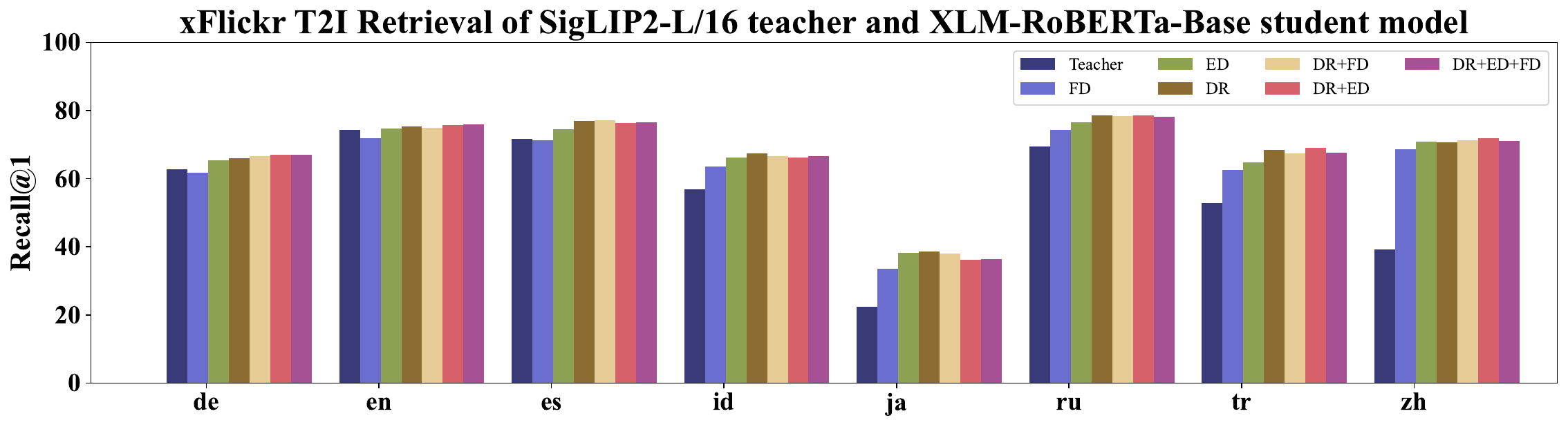}
    \vspace{-8mm}
    \caption{Result of recall@1 on the xFlickr dataset.}
    \vspace{-3mm}
    \label{fig:xFlickr_Time}
\end{figure*}

We demonstrate the full retrieval result of each language on benchmarks in Figures~\ref{fig:Multi30kI2T_Time}, \ref{fig:Multi30kT2I_Time}, \ref{fig:MSCOCOI2T_Time}, \ref{fig:MSCOCOT2I_Time}, \ref{fig:WITTime}, and \ref{fig:xFlickr_Time}.

\end{document}